\documentclass[sigconf]{acmart}
\AtBeginDocument{%
  }

\copyrightyear{2025}
\acmYear{2025}
\setcopyright{cc}
\setcctype{by}
\acmConference[CIKM '25]{Proceedings of the 34th ACM International Conference on Information and Knowledge Management}{November 10--14, 2025}{Seoul, Republic of Korea}
\acmBooktitle{Proceedings of the 34th ACM International Conference on Information and Knowledge Management (CIKM '25), November 10--14, 2025, Seoul, Republic of Korea}\acmDOI{10.1145/3746252.3761660}
\acmISBN{979-8-4007-2040-6/2025/11}

\usepackage{booktabs}
\usepackage{multirow}

\begin{document}

\title{Towards Rational Pesticide Design with Graph Machine Learning Models for Ecotoxicology}

\author{Jakub Adamczyk}
\email{jadamczy@agh.edu.pl}
\orcid{0000-0003-4336-4288}
\affiliation{%
  \institution{Faculty of Computer Science, AGH University of Krakow}
  \city{Cracow}
  \country{Poland}
}

\thanks{Supervised by Witold Dzwinel (dzwinel@agh.edu.pl) and Wojciech Czech (czech@agh.edu.pl), AGH University of Krakow}

\begin{abstract}
This research focuses on rational pesticide design, using graph machine learning to accelerate the development of safer, eco-friendly agrochemicals, inspired by \textit{in silico} methods in drug discovery. With an emphasis on ecotoxicology, the initial contributions include the creation of ApisTox, the largest curated dataset on pesticide toxicity to honey bees. We conducted a broad evaluation of machine learning (ML) models for molecular graph classification, including molecular fingerprints, graph kernels, GNNs, and pretrained transformers. The results show that methods successful in medicinal chemistry often fail to generalize to agrochemicals, underscoring the need for domain-specific models and benchmarks. Future work will focus on developing a comprehensive benchmarking suite and designing ML models tailored to the unique challenges of pesticide discovery.
\end{abstract}




\maketitle

\section{Introduction}

Pesticides are a central part of agrochemistry, comprising herbicides, insecticides, fungicides, and more. They are indispensable for modern agriculture and crop efficiency. However, growing concern surrounds their safety, both for humans (e.g. carcinogenicity, bioaccumulation) and the environment, particularly toxicity to honey bees, fish, birds, and small mammals. Pesticides must be bioactive, often targeting proteins similarly to drugs but with toxic, pest-killing effects. There is often a tradeoff between effectiveness and selectivity, i.e. targeting only specific organism groups, such as parasitic fungi. Computationally, this results in a complex multimodal optimization problem: we seek high lethality to pests, low toxicity to many beneficial (and often evolutionarily distant) species, long-lasting crop protection, but low bioaccumulation, etc.

Agrochemicals, like drugs, are small compounds (typically less than 50 atoms), represented as molecular graphs consisting of atoms and bonds, i.e. attributed graphs. The idea of rational drug design \cite{rational_drug_design}, where compound development is guided by computational predictions, has transformed the pharmaceutical industry. \textit{In silico} methods based on data mining and machine learning (ML) estimate key properties of candidate molecules, e.g. solubility, absorption, and toxicity, using prior experimental data. These methods help prioritize promising compounds for expensive wet lab testing, accelerating discovery and cutting costs.

In contrast, data science in agrochemistry remains nascent, and the field is still highly reliant on costly, time-consuming laboratory and field measurements. These deficiencies have consequences: registering a new active agrochemical costs over \$300M and takes about 12 years \cite{pesticide_cost}. Meanwhile, the EU Farm-to-Fork strategy aims to cut the use and risk of chemical pesticides by 50\% by 2030 \cite{eu_farm_to_fork}, driving the demand for faster development of safer alternatives. Laboratory studies are also ethically constrained, as ecotoxicity is commonly assessed using LD50, the dose that kills 50\% test organisms, such as honey bees or rats.

In my PhD work, I propose the concept of \textbf{rational pesticide design}, leveraging advances in graph machine learning and molecular data mining for novel agrochemical design. From a data science viewpoint, pesticides can be treated similarly to small drug-like molecules, with many of the same chemoinformatics tools applicable. The major challenge is the scarcity of publicly available datasets and benchmarks. As a result, the real-world performance of most molecular graph classification algorithms outside of medicinal chemistry remains unknown. This hinders fair evaluation, as most models are tested on only a few datasets from standard benchmarks, such as MoleculeNet \cite{MoleculeNet}.

The research questions in my work are as follows:

\noindent
\textbf{RQ1:} How can we design a reproducible data processing pipeline to generate high-quality datasets for pesticide property prediction?

\noindent
\textbf{RQ2:} How well do molecular graph classification algorithms perform in the agrochemical domain, especially those achieving state-of-the-art (SOTA) results in medicinal chemistry?

\noindent
\textbf{RQ3:} Are agrochemical datasets sufficiently distinct and challenging from medicinal chemistry to serve as meaningful benchmarks for molecular property prediction?

Initial experiments indicate that ML-based rational pesticide design is both feasible and effective, but introduces new challenges. The results also question the generalizability of methods that achieve SOTA performance on medicinal datasets, highlighting the need for agrochemistry-specific benchmarks.

\section{State of the art}

\noindent
\textbf{Agrochemistry.} The use of ML in agrochemistry remains very limited. The few available datasets focus almost exclusively on ecotoxicology - the most regulated aspect of pesticide design - overseen by agencies such as the US EPA and EU EFSA. Honey bees are of particular interest, as both vital pollinators and economically significant organisms \cite{ApisTox,ApisToxML,EPA_guidelines}.

Existing datasets, such as CropCSM \cite{CropCSM} and BeeTOX \cite{BeeTox}, are small, often just a few hundred molecules, and suffer from quality issues like invalid structures and duplicates \cite{ApisTox}. Data curation in agrochemistry is especially challenging and requires domain-specific approaches. For instance, while salts, inorganics, mixtures, and organometallics are typically excluded in medicinal chemistry, they often carry crucial information in pesticides \cite{EPA_QSAR_guidelines}.

Measurement errors are significant in toxicology, especially in ecotoxicology - a challenge acknowledged by the US EPA \cite{EPA_guidelines}. Regulators define official LD50 toxicity thresholds, allowing the molecular graph regression task to be framed as binary classification. For example, pesticides with LD50 below 11 $\mu$g/organism for honey bees are classified as highly toxic (positive class). Similar thresholds apply to other species, such as rats \cite{ld50_rats} or fish \cite{ld50_fish}.

\noindent
\textbf{Graph classification.} Many data mining approaches to graph classification have been created over the years. Simple baselines, utilizing topological graph descriptors or distributions of atoms and bonds, have shown remarkable performance in many cases, e.g. LTP \cite{LTP} and MOLTOP \cite{MOLTOP}. Graph kernels provide a flexible way to create a pairwise molecule similarity matrix, coupled with SVM classifier, e.g. Weisfeiler-Lehman (WL) or WL Optimal Assignment (WL-OA) kernels \cite{graph_kernels}.

Molecular fingerprints are a staple of chemoinformatics, providing automated feature extraction from molecular graphs \cite{scikit_fingerprints}. They fall into two main types: substructural fingerprints, which use predefined patterns designed by domain experts (e.g., MACCS, Laggner), and hashed fingerprints, which encode all subgraph occurrences of a given shape, such as circular neighborhoods (ECFP) or short paths (Topological Torsion). These representations are typically used with tree-based ensemble methods for classification.

Graph neural networks (GNNs) are neural networks with graph-based inductive biases, such as the permutation invariance of graph nodes and edges \cite{GNNs_survey}. They follow a message-passing paradigm, where graph convolution combines the feature vector of each atom with those of its neighbors at each layer. GNNs vary in their aggregation mechanisms, e.g. GCN, GraphSAGE, GIN, and GAT. Several architectures are tailored for molecular graphs, such as AttentiveFP. These models learn task-specific features and are typically trained from scratch.

Many approaches have been proposed to pretrain molecular ML models. GNN-based graph transformers such as MAT \cite{pretrained_MAT}, R-MAT \cite{pretrained_RMAT}, and GROVER \cite{pretrained_GROVER} combine the inductive biases of GNNs with the high expressiveness of transformers. Alternatively, molecular graphs can be serialized as SMILES strings - an efficient format for datasets and text-based data mining. NLP-inspired models, such as the ChemBERTa transformer \cite{pretrained_ChemBERTa2} and Mol2Vec \cite{mol2vec} (which combines ECFP fingerprints with Word2Vec), operate directly on SMILES. Given the small size of molecular datasets, to avoid overfitting during finetuning, embeddings from these models are often used as pretrained feature extractors.

\begin{figure}[H]
    \centering
    \includegraphics[width=0.4\textwidth]{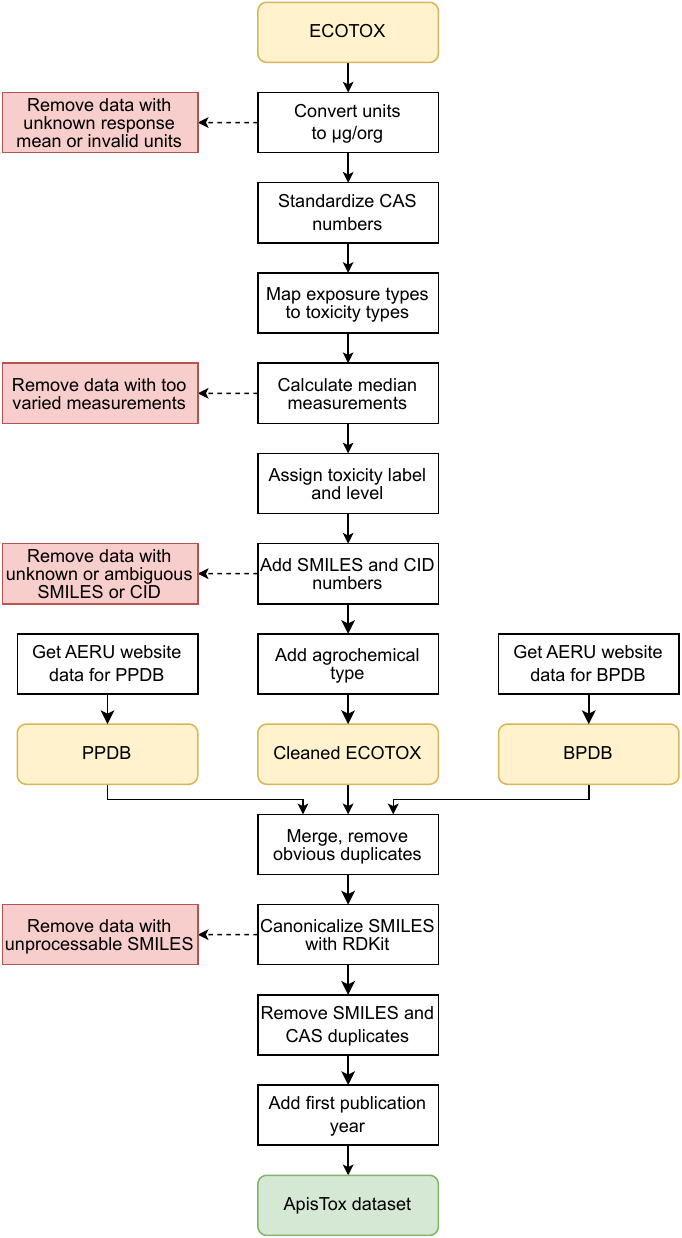}
    \caption{ApisTox data processing workflow \cite{ApisTox}.}
    \label{workflow}
\end{figure}

\section{Approach and methodology}

\textbf{ApisTox dataset.} As a practical first step in rational pesticide design, we focused on ecotoxicity, specifically for honey bees (\textit{Apis mellifera}). Three major public databases cover this area: ECOTOX, PPDB, and BPDB. ECOTOX contains individual experimental measurements, but requires extensive curation. The pipeline shown in Figure \ref{workflow} is designed to be general and applies to most ecotoxicology endpoints in ECOTOX. Due to space constraints, only a brief overview is provided here; full details are given in \cite{ApisTox}.

First, all units are standardized to $\mu$g/organism. For each pesticide, measurements are grouped by toxicity type - oral, contact, or other - and the median value per group is taken for. The lowest of the three medians (i.e. the strongest toxicity) is used as the overall LD50. SMILES strings are added using CAS numbers via the PubChem database.

PPDB and BPDB, which are manually curated and provide one record per pesticide, are merged with the preprocessed ECOTOX data. Molecular structures are standardized with RDKit, ensuring no structural duplicates (i.e. identical molecular graphs). Additional metadata include pesticide type (e.g. herbicide) and the first publication date of the literature, allowing for more detailed data analysis.

Importantly, this pipeline is reusable for other ecotoxicology endpoints, for e.g. for algae, fish, or birds. The user only needs to specify the toxicity threshold for the organism of interest.

\textbf{Predictive ML models.} To evaluate the usefulness of ApisTox, several molecular graph classification algorithms were implemented. Baselines included simple atom counts, LTP, and MOLTOP. Over 30 molecular fingerprints were generated using scikit-fingerprints \cite{scikit_fingerprints}, with Random Forest as a classifier. Graph kernels included top-performing options from the literature, particularly WL and WL-OA kernels. GNNs covered general-purpose models - GCN, GraphSAGE, GIN, and GAT - as well as chemistry-specific AttentiveFP. All were trained from scratch with extensive hyperparameter tuning. Pretrained models included graph- and SMILES-based architectures: MAT, R-MAT, GROVER, ChemBERTa, and Mol2Vec. These were used as feature extractors without finetuning, which caused immediate overfitting in preliminary tests. Logistic regression was chosen as the classifier for pretrained embeddings, as it yielded slightly better and more stable results than Random Forest.

\textbf{Performance evaluation.} Random train-test split in molecular data tend to overestimate model performance, as nearly identical molecules often appear in both sets \cite{MoleculeNet}. In medicinal chemistry, scaffold split is used to address this, utilizing the internal graph structure, but it does not work for salts (disconnected graphs), which are common in agrochemistry. We propose two novel approaches: MaxMin split and time split. MaxMin split selects the test molecules to maximize their total distance, creating a diverse test set that uniformly covers the chemical space. The time split assigns the newest molecules to the test set, based on their publication dates in the literature, similar to the real-world pesticide design process.

\textbf{Molecular diversity.} A common measure of molecular diversity in chemoinformatics is the average pairwise Tanimoto similarity between ECFP4 fingerprints \cite{tanimoto_similarity}. This metric quantifies both inter- and intra-dataset diversity, enabling meaningful comparisons between datasets and benchmarks. It was applied here to compare ApisTox with MoleculeNet classification datasets.

\section{Results}

\begin{table}[t]
\caption{Classification results. The best metric value for each split (column), in each group, is marked in bold.}
\label{table_results}
\resizebox{0.4\textwidth}{!}{
\begin{tabular}{@{}|c|c|cc|@{}}
\toprule
\multirow{2}{*}{\textbf{Group}} & \multirow{2}{*}{\textbf{Method}} & \multicolumn{2}{c|}{\textbf{MCC}} \\ \cmidrule(l){3-4} 
 &  & \multicolumn{1}{c|}{\textbf{MaxMin split}} & \textbf{Time split} \\ \midrule
\multirow{5}{*}{Fingerprints} & Atom Pairs & \multicolumn{1}{c|}{0.45 $\pm$ 0.03} & 0.37 $\pm$ 0.03 \\ \cmidrule(l){2-4} 
 & Avalon & \multicolumn{1}{c|}{\textbf{0.48 $\pm$ 0.03}} & 0.43 $\pm$ 0.02 \\ \cmidrule(l){2-4} 
 & ECFP & \multicolumn{1}{c|}{0.42 $\pm$ 0.02} & \textbf{0.48 $\pm$ 0.02} \\ \cmidrule(l){2-4} 
 & RDKit & \multicolumn{1}{c|}{0.43 $\pm$ 0.03} & 0.46 $\pm$ 0.02 \\ \cmidrule(l){2-4} 
 & Laggner & \multicolumn{1}{c|}{0.46 $\pm$ 0.03} & 0.37 $\pm$ 0.03 \\ \midrule
\multirow{3}{*}{Baselines} & Atom counts & \multicolumn{1}{c|}{\textbf{0.36 $\pm$ 0.03}} & 0.29 $\pm$ 0.04 \\ \cmidrule(l){2-4} 
 & LTP & \multicolumn{1}{c|}{0.18 $\pm$ 0.02} & 0.23 $\pm$ 0.01 \\ \cmidrule(l){2-4} 
 & MOLTOP & \multicolumn{1}{c|}{\textbf{0.36 $\pm$ 0.03}} & \textbf{0.33 $\pm$ 0.01} \\ \midrule
\multirow{4}{*}{\begin{tabular}[c]{@{}c@{}}Graph\\ kernels\end{tabular}} & Propagation & \multicolumn{1}{c|}{0.32} & 0.36 \\ \cmidrule(l){2-4} 
 & Shortest paths & \multicolumn{1}{c|}{0.29} & 0.31 \\ \cmidrule(l){2-4} 
 & WL & \multicolumn{1}{c|}{0.42} & 0.41 \\ \cmidrule(l){2-4} 
 & WL-OA & \multicolumn{1}{c|}{\textbf{0.49}} & \textbf{0.43} \\ \midrule
\multirow{5}{*}{GNNs} & GCN & \multicolumn{1}{c|}{0.25 $\pm$ 0.04} & 0.30 $\pm$ 0.04 \\ \cmidrule(l){2-4} 
 & GraphSAGE & \multicolumn{1}{c|}{0.31 $\pm$ 0.05} & \textbf{0.33 $\pm$ 0.04} \\ \cmidrule(l){2-4} 
 & GIN & \multicolumn{1}{c|}{0.24 $\pm$ 0.04} & 0.32 $\pm$ 0.06 \\ \cmidrule(l){2-4} 
 & GAT & \multicolumn{1}{c|}{0.26 $\pm$ 0.03} & 0.26 $\pm$ 0.05 \\ \cmidrule(l){2-4} 
 & AttentiveFP & \multicolumn{1}{c|}{\textbf{0.35 $\pm$ 0.04}} & 0.29 $\pm$ 0.06 \\ \midrule
\multirow{5}{*}{\begin{tabular}[c]{@{}c@{}}Pretrained\\ neural\\ networks\end{tabular}} & MAT & \multicolumn{1}{c|}{0.36} & 0.25 \\ \cmidrule(l){2-4} 
 & R-MAT & \multicolumn{1}{c|}{0.31} & \textbf{0.35} \\ \cmidrule(l){2-4} 
 & GROVER & \multicolumn{1}{c|}{0.22} & 0.05 \\ \cmidrule(l){2-4} 
 & ChemBERTa & \multicolumn{1}{c|}{\textbf{0.37}} & 0.27 \\ \cmidrule(l){2-4} 
 & Mol2Vec & \multicolumn{1}{c|}{0.34} & 0.31 \\ \bottomrule
\end{tabular}
}
\end{table}

Using the designed workflow, we created the ApisTox dataset \cite{ApisTox}. It consists of 1035 pesticide molecules in SMILES format with binary toxic/non-toxic labels, following US EPA guidelines. It is moderately imbalanced, with 29\% positive cases. Compared to previous datasets (e.g., CropCSM, BeeTox), it is the largest and the only one free of invalid entries or structural duplicates.

The initial classification results are presented in Table \ref{table_results}, with more details in the preprint \cite{ApisToxML}. Five top-performing molecular fingerprints were selected for brevity. Matthews correlation coefficient (MCC) was used as an evaluation metric, as it works well for imbalanced classification.

The main observation is that ApisTox is challenging, with the best method achieving an MCC of 0.48. In both splits, the best method is a molecular fingerprint, calling into question the effectiveness of GNNs on small datasets typical of agrochemistry. In fact, all GNNs and nearly all pretrained neural networks fail to outperform the MOLTOP baseline. Graph kernels, particularly the WL-OA kernel, perform strongly, achieving the best results on the MaxMin split. These findings differ significantly from medicinal chemistry, so how can they be explained?

\begin{figure}[t]
    \centering
    \includegraphics[width=0.5\textwidth]{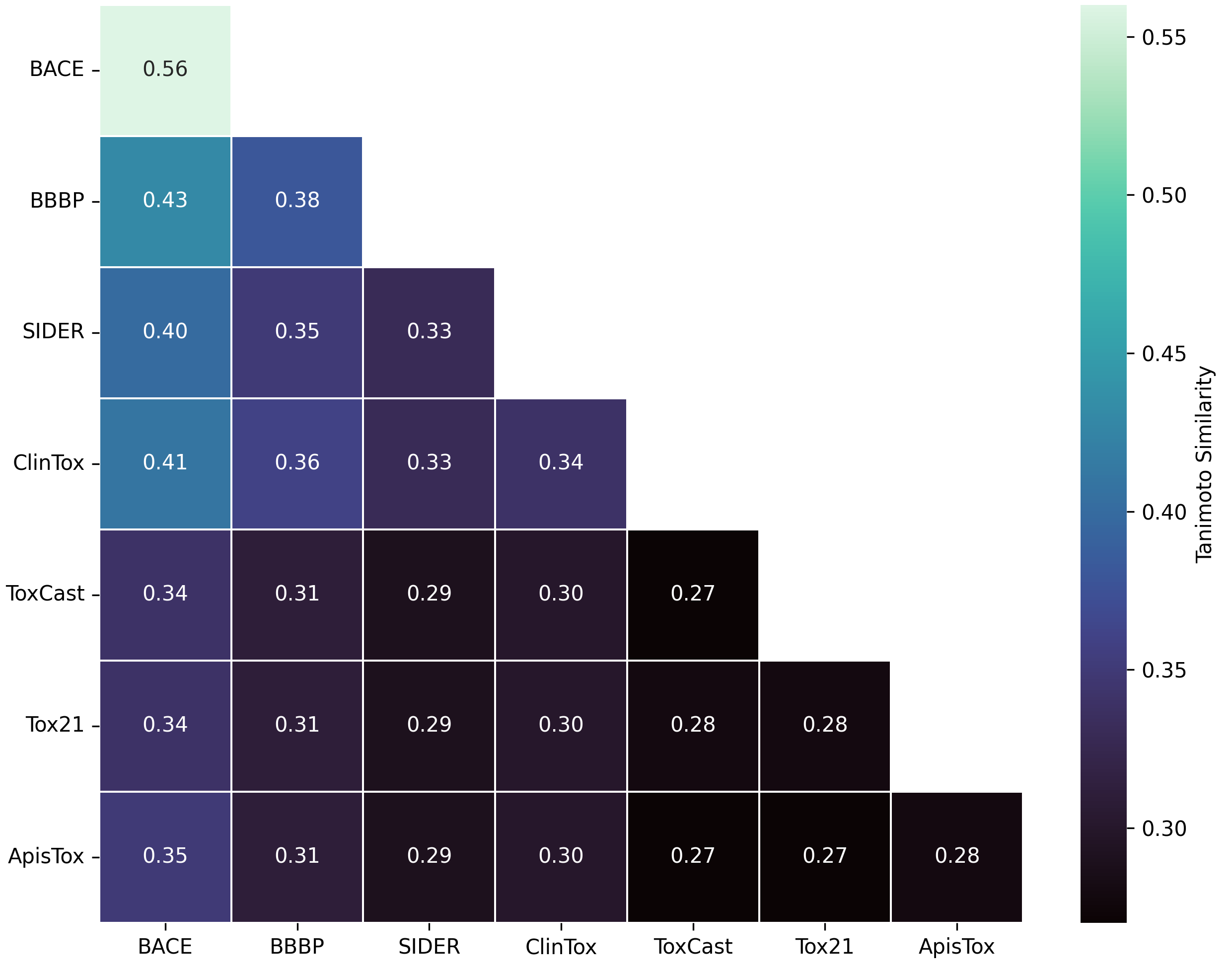}
    \caption{Average Tanimoto similarity between molecules from different datasets.}
    \label{tanimito_sim_map}
\end{figure}

First, pretrained models often rely on heavily filtered medicinal data. For example, MAT and R-MAT were pretrained on the ZINC dataset, further filtered by the conservative Lipinski Rule of 5. This inherently biases the models toward a limited data distribution. Additionally, a quantitative comparison of ApisTox and the commonly used MoleculeNet datasets (see Figure \ref{tanimito_sim_map}) shows that ApisTox is highly distinct from typical medicinal compounds. Meanwhile, MoleculeNet is internally homogeneous; for instance, BACE and BBBP datasets show high internal similarity. This limits its usefulness, as models are implicitly encouraged to overfit these common patterns to boost benchmark scores.

\section{Conclusion and future work}

Machine learning, graph mining, and predictive analytics are well established in pharmaceutical sciences and drug design, yet their application in agrochemistry remains nascent. This work introduces the concept of \textbf{rational pesticide design}, applying computational \textit{in silico} approaches to aid the development of new agrochemicals. By mirroring pharmaceutical computational methods, we aim to reduce costs and accelerate novel agrochemical development.

Initial efforts include creating the ApisTox dataset \cite{ApisTox} of pesticide toxicity to honey bees and training predictive ML models for molecular graph classification \cite{ApisToxML}. The results show that agrochemistry occupies a chemical space distinct from those of medicinal chemistry benchmarks such as MoleculeNet. Furthermore, models pretrained on pharmaceutical data appear overtuned, failing to generalize to novel agrochemical molecules. This highlights the need for new datasets and benchmarks beyond medicinal chemistry, providing initial answers to the research questions RQ1, RQ2, and RQ3, although further work is required for definitive conclusions.

Future work includes creating a comprehensive pesticides benchmark to fairly evaluate molecular graph classification models in this chemical space. We will also develop agrochemistry-specific predictive models, supported by large-scale pretraining datasets with reduced bias toward medicinal compounds.




\begin{acks}
I thank my supervisors, Prof. Witold Dzwinel and Dr. Wojciech Czech. I thank Jakub Poziemski and Paweł Siedlecki, coauthors of ApisTox and subsequent ML modeling for this dataset. I thank BIT Student Scientific Group for computational resources and aid for this project. This work was supported by the Ministry of Science and Higher Education funds assigned to AGH University in Krakow. I thank Aleksandra Elbakyan for her work and support for accessibility of science.
\end{acks}

\bibliographystyle{ACM-Reference-Format}
\bibliography{bibliography}

\end{document}